\pdfoutput=1
\documentclass[11pt]{article}

\usepackage{acl}

\usepackage{times}
\usepackage{latexsym}

\usepackage[T1]{fontenc}
\usepackage[utf8]{inputenc}

\usepackage{multirow}
\usepackage{makecell}
\usepackage{booktabs} % for professional tables

\usepackage{amsmath}
\usepackage{amssymb}
\usepackage{amsfonts}       % blackboard math symbols
\usepackage{nicefrac}       % compact symbols for 1/2, etc.
\usepackage{microtype}
\usepackage{xcolor}         % colors

\usepackage{hyperref}
\usepackage{cleveref}
\usepackage{url}            % simple URL typesetting

\usepackage{graphicx}
\usepackage{caption}
\usepackage{subcaption}
\usepackage{tikz}

\usepackage{amsthm}
\theoremstyle{definition}
\newtheorem{example}{Example}[section]

\usepackage{xspace}

\newcommand{\model}{GLM\xspace}
\newcommand{\modelx}{GLM}

\newcommand{\myvec}[1]{\boldsymbol{#1}}

\newcommand{\E}{\mathbb{E}}
\newcommand{\base}{$_\text{Base}$\xspace}
\newcommand{\largem}{$_\text{Large}$\xspace}
\newcommand{\largemulti}{$_\text{Doc}$\xspace}
\newcommand{\roberta}{$_{\text{RoBERTa}}$\xspace}
\newcommand{\largemodel}{GLM$_\text{Large}$\xspace}
\newcommand{\docmodel}{GLM$_\text{Doc}$\xspace}
\newcommand{\sentmodel}{GLM$_\text{Sent}$\xspace}

\title{GLM: General Language Model Pretraining \\ with Autoregressive Blank Infilling}

\author{
  Zhengxiao Du$^{*1,2}$ \quad Yujie Qian$^{*3}$ \quad Xiao Liu$^{1,2}$ \quad Ming Ding$^{1,2}$ \quad Jiezhong Qiu$^{1,2}$\\ \bf Zhilin Yang\footnotemark[2]$\;^{1,4}$ \quad Jie Tang\footnotemark[2]$\;^{1,2}$\\
  $^1$Tsinghua University \quad $^2$Beijing Academy of Artificial Intelligence (BAAI)\\
  $^3$MIT CSAIL \quad $^4$Shanghai Qi Zhi Institute\\
  \texttt{zx-du20@mails.tsinghua.edu.cn}\quad \texttt{yujieq@csail.mit.edu}\\
  \texttt{\{zhiliny,jietang\}@tsinghua.edu.cn}
}

\begin{document}
\maketitle
\renewcommand{\thefootnote}{\fnsymbol{footnote}}
\footnotetext[1]{The first two authors contributed equally.}
\footnotetext[2]{Corresponding authors.}
\renewcommand{\thefootnote}{\arabic{footnote}}

\begin{abstract}
  There have been various types of pretraining architectures including autoencoding models (e.g., BERT), autoregressive models (e.g., GPT), and encoder-decoder models (e.g., T5). However, none of the pretraining frameworks performs the best for all tasks of three main categories including natural language understanding (NLU), unconditional generation, and conditional generation. We propose a General Language Model (\model)  based on autoregressive blank infilling to address this challenge.
  \model improves blank filling pretraining by adding 2D positional encodings and allowing an arbitrary order to predict spans, which results in performance gains over BERT and T5 on NLU tasks. 
%   Empirically, \model is flexible enough to perform competetively on various NLP tasks with a single pretrained model. 
%   The proposed architecture has two major benefits: (1) It improves pretrain-finetune consistency via cloze-style finetuning and naturally handles variable-length blank infilling which is crucial for many downstream tasks. Empirically, \model substantially outperforms BERT on the SuperGLUE natural language understanding benchmark with the same amount of pretraining data and steps. (2) It is flexible enough to handle various NLP tasks with a single pretrained model. 
Meanwhile, \model can be pretrained for different types of tasks by varying the number and lengths of blanks.
On a wide range of tasks across NLU, conditional and unconditional generation, \model outperforms BERT, T5, and GPT given the same model sizes and data, and achieves the best performance from a single pretrained model with 1.25$\times$ parameters of BERT\largem, demonstrating its generalizability to different downstream tasks.\footnote{The code and pre-trained models are available at \url{https://github.com/THUDM/GLM}}
\end{abstract}

\section{Introduction}
\label{sec:intro}
Language models pretrained on unlabeled texts have substantially advanced the state of the art in various NLP tasks, ranging from natural language understanding (NLU) to text generation~\cite{GPT2018a,devlinBERT2019,YangXLNet2019,GPT2-2018,raffelT52020,lewisBART2019,GPT3-2020}. % The scale of the largest pretrained models has increased by over 1,000 times in the past two years.
Downstream task performance as well as the scale of the parameters have also constantly increased in the past few years.

\begin{figure}
    \centering
    \includegraphics[width=\linewidth]{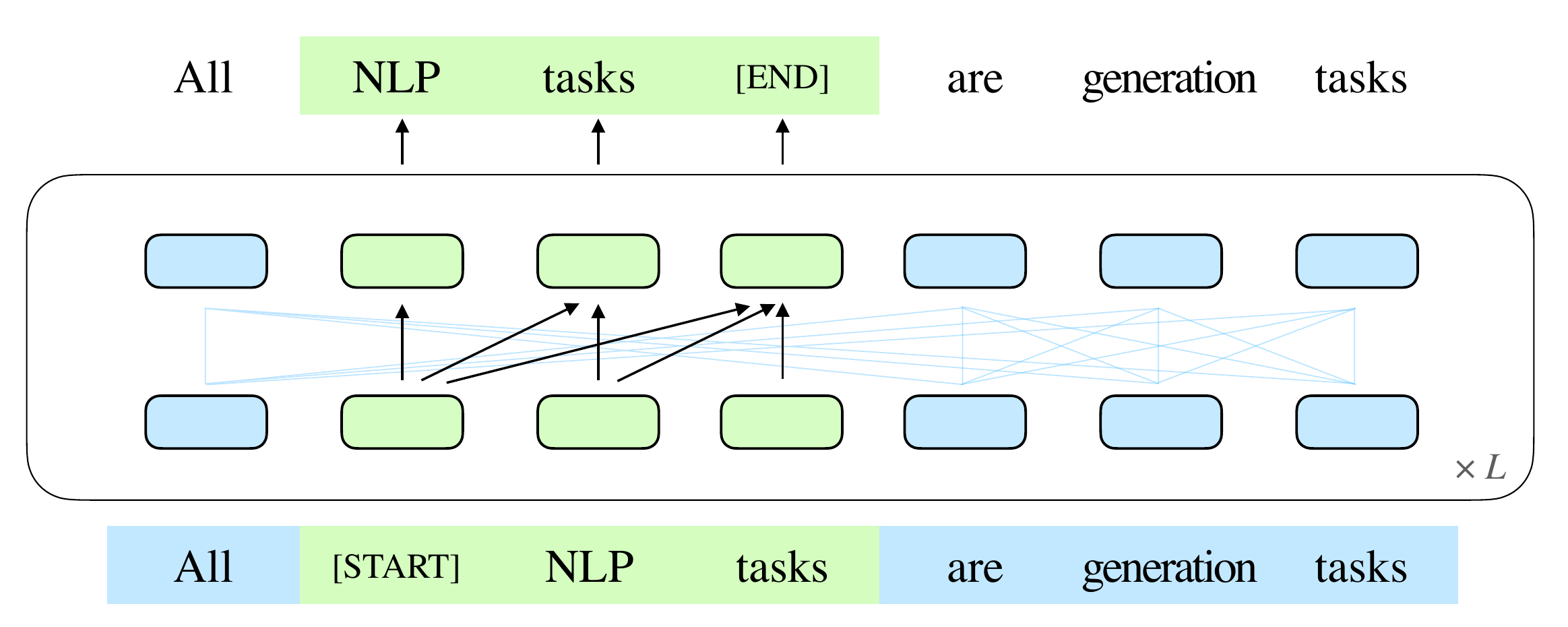}
    \caption{Illustration of \model. We blank out text spans (green part) and generate them autoregressively. (Some attention edges are omitted; cf. Figure~\ref{fig:model}.)}
    \label{fig:example}
\end{figure}

In general, existing pretraining frameworks can be categorized into three families: \textit{autoregressive}, \textit{autoencoding}, and \textit{encoder-decoder} models. Autoregressive models, such as GPT~\cite{GPT2018a}, learn left-to-right language models. While they succeed in long-text generation and show few-shot learning ability when scaled to billions of parameters~\cite{GPT2-2018,GPT3-2020}, the inherent disadvantage is the unidirectional attention mechanism, which cannot fully capture the dependencies between the context words in NLU tasks. Autoencoding models, such as BERT~\cite{devlinBERT2019}, learn bidirectional context encoders via denoising objectives, e.g. Masked Language Model (MLM). The encoders 
produce contextualized representations that suit natural language understanding tasks, but could not be directly applied for text generation. Encoder-decoder models adopt bidirectional attention for the encoder, unidirectional attention for the decoder, and cross attention between them~\cite{songMASS2019,PALM2020,lewisBART2019}. They are typically deployed in conditional generation tasks, such as text summarization and response generation.
\footnote{Unconditional generation refers to generating text as a language model without finetuning, while conditional generation refers to sequence-to-sequence tasks.}. 
T5~\cite{raffelT52020} unifies NLU and conditional generation via encoder-decoder models but requires more parameters to match the performance of BRET-based models such as RoBERTa~\cite{RoBERTa} and DeBERTa \cite{He2021DeBERTaDB}. 
% \Cref{tab:summary} compares different pretraining frameworks.

None of these pretraining frameworks is flexible enough to perform competitively across all NLP tasks. %Considering the high computational cost of pretraining a large language model, it hinders downstream task development and model selection. 
Previous works have tried to unify different frameworks by combining their objectives via multi-task learning~\cite{UniLM19,UniLMv220}. However, since the autoencoding and autoregressive objectives differ by nature, a simple unification cannot fully inherit the advantages of both frameworks.

In this paper, we propose a pretraining framework named \model (General Language Model), based on autoregressive blank infilling. We randomly blank out continuous spans of tokens from the input text, following the idea of autoencoding, and train the model to sequentially reconstruct the spans, following the idea of autoregressive pretraining (see \Cref{fig:example}). While blanking filling has been used in T5~\cite{raffelT52020} for text-to-text pretraining, we propose two improvements, namely span shuffling and 2D positional encoding. Empirically, we show that with the same amount of parameters and computational cost, \model significantly outperforms BERT on the SuperGLUE benchmark by a large margin of 4.6\% -- 5.0\% and outperforms RoBERTa and BART when pretrained on a corpus of similar size (158GB). GLM also significantly outperforms T5 on NLU and generation tasks with fewer parameters and data.
% \zy{add comparison with T5 on seq2seq tasks}
% To learn both bidirectional and unidirectional attention mechanisms in a unified framework, we divide the input text into two parts, where the unmasked tokens can attend to each other, but the masked tokens cannot attend to subsequent masked tokens. 
% Figure~\ref{fig:model} illustrates our pretraining framework. 

Inspired by Pattern-Exploiting Training (PET) \cite{PET:2001-07676}, we reformulate NLU tasks as manually-crafted cloze questions that mimic human language. Different from the BERT-based models used by PET, GLM can naturally handle multi-token answers to the cloze question via autoregressive blank filling.
% When finetuning on downstream tasks, we reformulate them as blank infilling generation, 
% For example, a sentiment classification task is reformulated as a cloze question ``[SENTENCE]. It's really \rule{0.4cm}{0.4pt}''. The choice of words ``good'' or ``bad'' in the blank indicates the sentiment being positive or negative. 
% With such formulation, \model benefits from the consistency between pretraining and finetuning, because they both train the model to generate masked spans given their contexts. 

Furthermore, we show that by varying the number and lengths of missing spans, the autoregressive blank filling objective can pretrain language models for conditional and unconditional generation. Through multi-task learning of different pretraining objectives, a single GLM can excel in both NLU and (conditional and unconditional) text generation.
Empirically, compared with standalone baselines, \model with multi-task pretraining achieves improvements in NLU, conditional text generation, and language modeling tasks altogether by sharing the parameters.

\begin{figure*}
    \centering
    \includegraphics[width=0.98\textwidth]{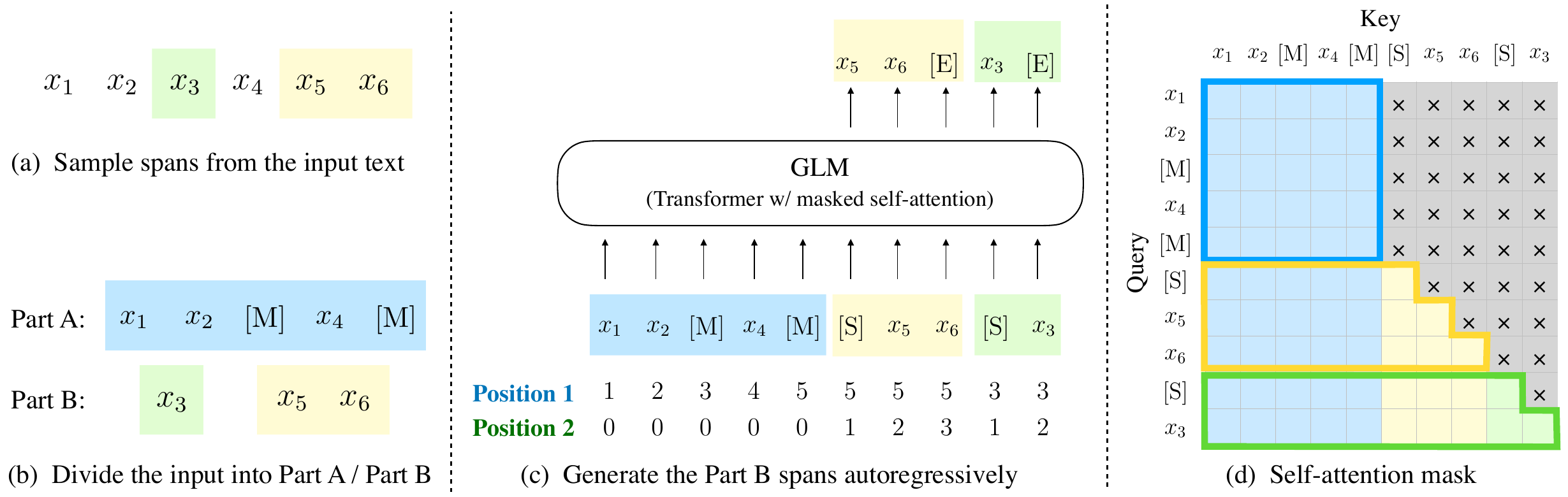}
    \caption{\model pretraining. (a) The original text is $[x_1,x_2,x_3,x_4,x_5,x_6]$. Two spans $[x_3]$ and $[x_5,x_6]$ are sampled. (b) Replace the sampled spans with [M] in Part A, and shuffle the spans in Part B. (c)  \model autoregressively generates Part B. Each span is prepended with [S] as input and appended with [E] as output. 2D positional encoding represents inter- and intra-span positions.  (d) Self-attention mask. Grey areas are masked out. Part A tokens can attend to themselves (blue frame) but not B. Part B tokens can attend to A and their antecedents in B (yellow and green frames correspond to the two spans).  $[\text{M}]:=[\text{MASK}]$, $[\text{S}]:=[\text{START}]$, and $[\text{E}]:=[\text{END}]$.}
    \label{fig:model}
\end{figure*}

\section{\model Pretraining Framework}
We propose a general pretraining framework GLM based on a novel autoregressive blank infilling objective. GLM formulates NLU tasks as cloze questions that contain task descriptions, which can be answered by autoregressive generation. 
%GLM can incorporate different NLP tasks by varying the number and lengths of blank spans. Specifically, GLM formulates the NLU tasks as cloze questions that contain some form of task descriptions, which can be solved by autoregressive blank infilling. 
% In this section, we introduce the pretraining objective, model architecture, and finetuning method of GLM.

\subsection{Pretraining Objective}
\label{subsec:objective}

\subsubsection{Autoregressive Blank Infilling}
\label{subsec:blankinfill}

\model is trained by optimizing an \textit{autoregressive blank infilling} objective. Given an input text $\myvec{x}=[x_1,\cdots,x_n]$, multiple text spans $\{\myvec{s}_1,\cdots,\myvec{s}_m\}$ are sampled, where each span $\myvec{s}_i$ corresponds to a series of consecutive tokens $[s_{i,1},\cdots,s_{i,l_i}]$ in $\myvec{x}$.  Each span is replaced with a single $[\text{MASK}]$ token, forming a corrupted text $\myvec{x}_{\text{corrupt}}$. The model predicts the missing tokens in the spans from the corrupted text in an autoregressive manner, which means when predicting the missing tokens in a span, the model has access to the corrupted text \emph{and} the previously predicted spans. To fully capture the interdependencies between different spans, we randomly permute the order of the spans, similar to the permutation language model~\cite{YangXLNet2019}.
Formally, let $Z_m$ be the set of all possible permutations of the length-$m$ index sequence $[1,2,\cdots,m]$, and $\myvec{s}_{\myvec{z}_{<i}}$ be $[\myvec{s}_{z_1},\cdots,\myvec{s}_{z_{i-1}}]$, we define the pretraining objective as
\begin{equation}
    \max_\theta \E_{\myvec{z}\sim Z_m}\left[\sum_{i=1}^m\log p_\theta(\myvec{s}_{z_i}|\myvec{x}_{\text{corrupt}},\myvec{s}_{\myvec{z}_{<i}})\right]
    \label{eqn:objective}
\end{equation}
% It differs from SpanBERT~\cite{joshiSpanBERT2020} in the sense that the number of missing tokens in a span is unknown to the model. \model predicts the missing tokens in an autoregressive manner. 
We always generate the tokens in each blank following a left-to-right order, i.e. the probability of generating the span $\myvec{s}_i$ is factorized as:
\begin{align}
\begin{split}
    &p_\theta(\myvec{s}_i|\myvec{x}_{\text{corrupt}},\myvec{s}_{\myvec{z}_{<i}})\\
    =&\prod_{j=1}^{l_i} p(s_{i,j}|\myvec{x}_{\text{corrupt}},\myvec{s}_{\myvec{z}_{<i}},\myvec{s}_{i,<j})
\end{split}
\end{align}
We implement the autoregressive blank infilling objective with the following techniques. The input $\myvec{x}$ is divided into two parts: Part A is the corrupted text $\myvec{x}_{\text{corrupt}}$, and Part B consists of the masked spans. Part A tokens can attend to each other, but cannot attend to any tokens in B. Part B tokens can attend to Part A and antecedents in B, but cannot attend to any subsequent tokens in B. To enable autoregressive generation, each span is padded with special tokens $[\text{START}]$ and $[\text{END}]$, for input and output respectively. In this way, our model automatically learns a bidirectional encoder (for Part A) and a unidirectional decoder (for Part B) in a unified model. The implementation of \model is illustrated in \Cref{fig:model}.

%The number and length of text spans depend on the pretraining objective, which is described in \Cref{subsec:objective}.
%As described in Section~\ref{subsec:blankinfill}, the pretraining objective of \model is defined as autoregressive generation of the masked spans. For NLU tasks, it 
% Previous pretraining works have shown optimal performance when masking 15\% of the original tokens~\cite{devlinBERT2019,YangXLNet2019,raffelT52020}. 
We randomly sample spans of length drawn from a Poisson distribution with $\lambda=3$. We repeatedly sample new spans until at least 15\% of the original tokens are masked. Empirically, we have found that the 15\% ratio is critical for good performance on downstream NLU tasks.

\subsubsection{Multi-Task Pretraining}
\label{subsec:multitask}

In the previous section, \model masks short spans and is suited for NLU tasks. However, we are interested in pretraining a single model that can handle both NLU and text generation. We then study a \textit{multi-task pretraining} setup, in which a second objective of generating longer text is jointly optimized with the blank infilling objective. We consider the following two objectives:
\begin{itemize}
    \item Document-level. We sample a single span whose length is sampled from a uniform distribution over 50\%--100\% of the original length. The objective aims for long text generation.
    % The span is masked with a [MASK] token and the model autoregressively generates it. %This objective is a mixture of GPT~\cite{GPT2018a} and Prefix LM discussed by~\cite{raffelT52020}.
    \item Sentence-level. We restrict that the masked spans must be full sentences. Multiple spans (sentences) are sampled to cover 15\% of the original tokens. This objective aims for seq2seq tasks whose predictions are often complete sentences or paragraphs. % This is inspired by the finding of \cite{PEGASUS:zhang2020} that pretraining Transformers to predict missing sentences is helpful to abstractive summarization.
\end{itemize}
Both new objectives are defined in the same way as the original objective, i.e. Eq.~\ref{eqn:objective}. The only difference is the number of spans and the span lengths.

% We can change the pretraining objective of \model by varying the distribution of the span lengths and numbers. We consider two types of pretraining objectives, one for language understanding and one for language generation. The first type is the BERT-style pretraining for natural language understanding. Following BERT, the missing tokens make up 15\% of the original tokens. Empirically we have found that the ratio is critical for good performance on downstream NLU tasks. Following BART~\cite{lewisBART2019},  The second type is the generation pretraining for seq2seq and language modeling. A single span that covers 50\%-100\% of the original tokens are sampled. The span length is sampled from a uniform distribution. The objective can be viewed as a mix of language modeling and seq2seq.

\subsection{Model Architecture}
\model uses a single Transformer with several modifications to the architecture: (1) we rearrange the order of layer normalization and the residual connection, which has been shown critical for large-scale language models to avoid numerical errors~\cite{Megatron-LM}; (2) we use a single linear layer for the output token prediction; (3) we replace ReLU activation functions with GeLUs~\cite{GeLU:HendrycksG16}.

% therefore the output position $i$ is defined as
% \begin{equation}
%     \myvec{p}_i=\operatorname{softmax}(\myvec{h}_i^L\mymat{W}_o)
% \end{equation}
% where $\mymat{W}_o\in\mathbb{R}^{d_h\times|\mathcal{V}|}$ and $|\mathcal{V}|$ is the vocabulary size.
% First, the input tokens $[x_1,x_2,\cdots,x_n]$ are projected into embedding vectors $\mymat{H}^0=[\myvec{h}^0_1,\myvec{h}^0_2,\cdots,\myvec{h}^0_n]$ via a learnable embedding table. Then $L$ transformer layers are applied to compute the hidden states of the tokens. Each layer consists of a multi-head self-attention layer and a position-wise fully connected feed-forward network. 
% Specifically, a self-attention head in layer $l$ is defined as
% \begin{equation}
%     \begin{split}
%         &\mymat{Q}^l=\mymat{H}^l\mymat{W}^l_Q,\mymat{K}^l=\mymat{H}^l\mymat{W}^l_K,\mymat{V}^l=\mymat{H}^l\mymat{W}^l_V\\
%         &\mymat{A}^l=\operatorname{softmax}\left(\frac{\mymat{Q}^l(\mymat{K}^l)^T}{\sqrt{d_k}}+\mymat{M}\right)\mymat{V}^l
%     \end{split}
% \end{equation}
% where $\mymat{W}^l_Q, \mymat{W}^l_K, \mymat{W}^l_V\in\mathbb{R}^{d_h\times d_k}$ are model parameters. $\mymat{M}\in\mathbb{R}^{n\times n}$ is the matrix of self-attention mask, ${M}_{ij}=0$ indicates that token $x_i$ is allowed to attend to token $x_j$ and ${M}_{ij}=-\infty$ indicates prevention.

\subsubsection{2D Positional Encoding}
One of the challenges of the autoregressive blank infilling task is how to encode the positional information. Transformers rely on positional encodings to inject the absolute and relative positions of the tokens. 
We propose 2D positional encodings to address the challenge. Specifically, each token is encoded with two positional ids. The first positional id represents the position in the corrupted text $\myvec{x}_{\text{corrupt}}$. For the masked spans, it is the position of the corresponding $[\text{MASK}]$ token. The second positional id represents the intra-span position. For tokens in Part A, their second positional ids are $0$. For tokens in Part B, they range from 1 to the length of the span. The two positional ids are projected into two vectors via learnable embedding tables, which are both added to the input token embeddings.

Our encoding method ensures that the model is not aware of the length of the masked span when reconstructing them. It is an important difference as compared to other models. For example, XLNet~\cite{YangXLNet2019} encodes the original position so that it can perceive the number of missing tokens, and SpanBERT~\cite{joshiSpanBERT2020} replaces the span with multiple [MASK] tokens and keeps the length unchanged. Our design fits downstream tasks as usually the length of the generated text is unknown beforehand.
% As a similar autoregressive model, XLNet encodes the original positions for tokens in Part A and B. As a result, the model can perceive the number of missing tokens in a span.

\begin{figure}
    \centering
    \includegraphics[width=0.45\textwidth]{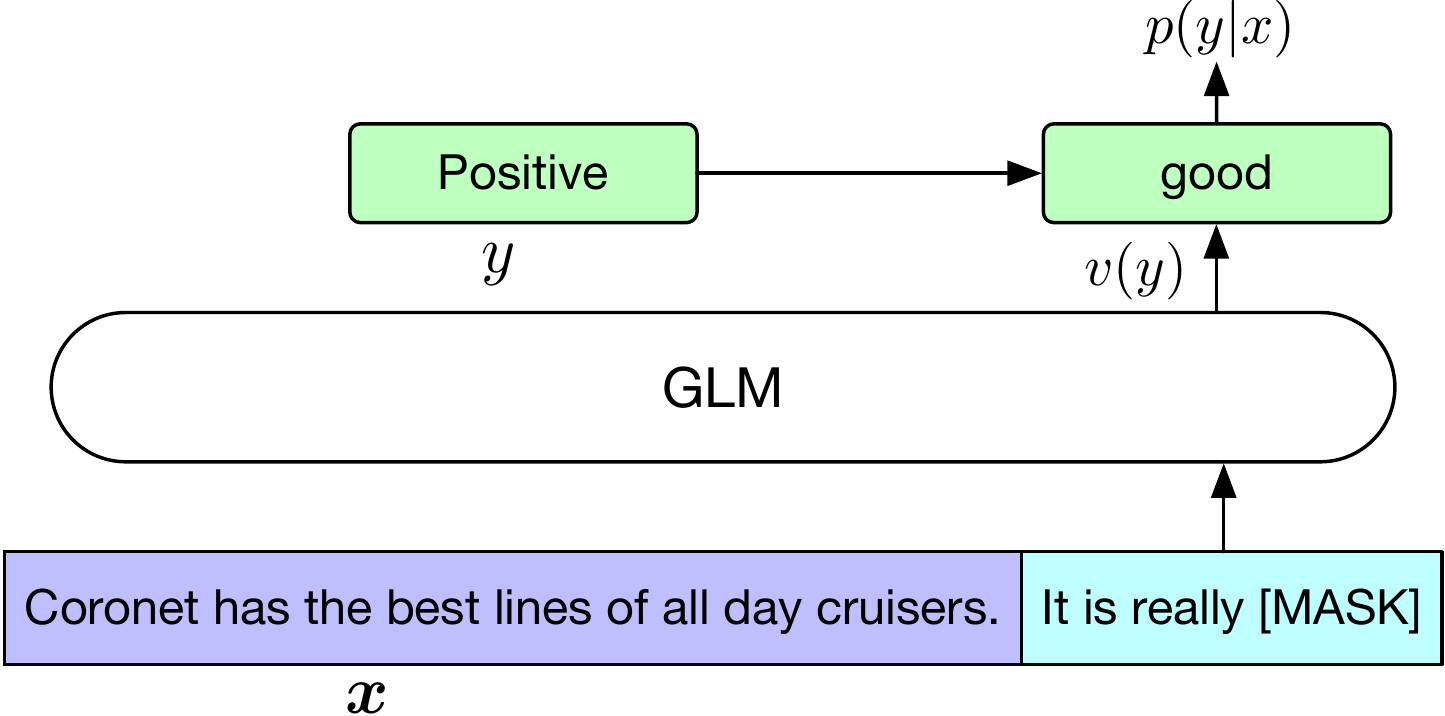}
    \caption{Formulation of the sentiment classification task as blank infilling with \model.}
    \label{fig:finetune}
\end{figure}

\subsection{Finetuning \model}
\label{subsec:finetune}
Typically, for downstream NLU tasks, a linear classifier takes the representations of sequences or tokens produced by pretrained models as input and predicts the correct labels.
% For token classification tasks, the inputs are representations of the target tokens. For sequence classification tasks, the input is the representation of the $[\text{CLS}]$ token, which works as the representation of the sequence. 
The practices are different from the generative pretraining task, leading to inconsistency between pretraining and finetuning. 

Instead, we reformulate NLU classification tasks as generation tasks of blank infilling, following PET~\cite{PET:2001-07676}. Specifically, given a labeled example $(\myvec{x}, y)$, we convert the input text $\myvec{x}$ to a cloze question $c(\myvec{x})$ via a pattern containing a single mask token. The pattern is written in natural language to represent the semantics of the task. For example, a sentiment classification task can be formulated as ``\{SENTENCE\}. It's really $[\text{MASK}]$''. The candidate labels $y\in\mathcal{Y}$ are also mapped to answers to the cloze, called verbalizer $v(y)$. In sentiment classification, the labels ``positive'' and ``negative'' are mapped to the words ``good'' and ``bad''. The conditional probability of predicting $y$ given $\myvec{x}$ is
\begin{equation}
    p(y|\myvec{x})=\frac{p(v(y)|c(\myvec{x}))}{\sum_{y'\in \mathcal{Y}}p(v(y')|c(\myvec{x}))}
\end{equation}
where $\mathcal{Y}$ is the label set. Therefore the probability of the sentence being positive or negative is proportional to predicting ``good'' or ``bad'' in the blank. Then we finetune \model with a cross-entropy loss (see \Cref{fig:finetune}).

% \model is especially suited to this setting for two reasons. Firstly, \model can naturally handle filling the blank of unknown length. BERT-style models have to know the number of missing tokens via the number of $[\text{MASK}]$ tokens or the positional encodings. Secondly, \model breaks BERT's independence assumption between masked tokens and thus can capture more dependencies.

For text generation tasks, the given context constitutes the Part A of the input, with a mask token appended at the end. The model generates the text of Part B autoregressively. We can directly apply the pretrained \model for unconditional generation, or finetune it on downstream conditional generation tasks.

\subsection{Discussion and Analysis}
In this section, we discuss the differences between \model and other pretraining models. % including BERT, XLNet, T5, and UniLM. Note that \model differs from the first three models in that \model incorporates the generative objective. 
We are mainly concerned with how they can be adapted to downstream blank infilling tasks.

\textbf{Comparison with BERT~\cite{devlinBERT2019}.}
% BERT is pretrained with an autoencoding objective, i.e. masked language model. 
% The model needs to predict a subset of tokens in the original text replaced with the $[\text{MASK}]$ token.
As pointed out by \cite{YangXLNet2019}, BERT fails to capture the interdependencies of masked tokens due to the independence assumption of MLM. Another disadvantage of BERT is that it cannot fill in the blanks of multiple tokens properly. To infer the probability of an answer of length $l$,  BERT needs to perform $l$ consecutive predictions. If the length $l$ is unknown, we may need to enumerate all possible lengths, since BERT needs to change the number of $[\text{MASK}]$ tokens according to the length.

\textbf{Comparison with XLNet~\cite{YangXLNet2019}.}
Both \model and XLNet are pretrained with autoregressive objectives, but there are two differences between them. First, XLNet uses the original position encodings before corruption. During inference, we need to either know or enumerate the length of the answer, the same problem as BERT. Second, XLNet uses a two-stream self-attention mechanism, instead of the right-shift, to avoid the information leak within Transformer. It doubles the time cost of pretraining.

\textbf{Comparison with T5~\cite{raffelT52020}.}
T5 proposes a similar blank infilling objective to pretrain an encoder-decoder Transformer. 
% Instead, \model uses a single Transformer encoder model to learn both bidirectional and unidirectional attention. By sharing parameters for the two types of attention, \model is more parameter-efficient than the encoder-decoder architecture. 
T5 uses independent positional encodings for the encoder and decoder, and relies on multiple sentinel tokens to differentiate the masked spans. In downstream tasks, only one of the sentinel tokens is used, leading to a waste of model capacity and inconsistency between pretraining and finetuning. Moreover, T5 always predicts spans in a fixed left-to-right order. As a result, \model can significantly outperform T5 on NLU and seq2seq tasks with fewer parameters and data, as stated in \Cref{subsec:single_obj,subsec:multi_obj}.

\textbf{Comparison with UniLM~\cite{UniLM19}.}
UniLM combines different pretraining objectives under the autoencoding framework by changing the attention mask among bidirectional, unidirectional, and cross attention. However, UniLM always replaces masked spans with [MASK] tokens, which limits its ability to model the dependencies between the masked spans and their context. \model feeds in the previous token and autoregressively generates the next token. Finetuning UniLM on downstream generation tasks also relies on masked language modeling, which is less efficient. UniLMv2~\cite{UniLMv220} adopts partially autoregressive modeling for generation tasks, along with the autoencoding objective for NLU tasks. Instead, \model unifies NLU and generation tasks with autoregressive pretraining.
\begin{table*}[t]
    \caption{Results on the SuperGLUE dev set.} %  Models with * are pretrained for two times the number of steps of other methods.
    % \vspace{-0.08in}
    % \small
    \label{tab:superglue}
    \centering
    \resizebox{0.98\textwidth}{!}{
    \begin{tabular}{ll*{10}{r}}
        \toprule
        &Model &  \makecell{ReCoRD\\F1/Acc.} & \makecell{COPA\\Acc.} & \makecell{WSC\\Acc.} & \makecell{RTE\\Acc.} & \makecell{BoolQ\\Acc.} &  \makecell{WiC\\Acc.} & \makecell{CB\\F1/Acc.}  & \makecell{MultiRC\\F1a/EM} & Avg \\
        \midrule\midrule
        \multicolumn{5}{l}{\textit{Pretrained on BookCorpus and Wikipedia}}\\[0.02in]
        &BERT\base & 65.4 / 64.9 & 66.0 & 65.4 & 70.0 & 74.9 & \textbf{68.8} & 70.9 / 76.8 & 68.4 / 21.5 & 66.1 \\
        &\modelx\base & \textbf{73.5 / 72.8} & \textbf{71.0} & \textbf{72.1} & \textbf{71.2} & \textbf{77.0} & 64.7 & \textbf{89.5 / 85.7} & \textbf{72.1 / 26.1} & \textbf{70.7} \\
        \midrule
        &BERT\largem & 76.3 / 75.6 & 69.0 & 64.4 & 73.6 & 80.1 & \textbf{71.0} & 94.8 / 92.9 & 71.9 / 24.1 & 72.0 \\
        &UniLM\largem & 80.0 / 79.1 & 72.0 & 65.4 & 76.5 & 80.5 & 69.7 & 91.0 / 91.1 & \textbf{77.2 / 38.2} & 74.1 \\
        &\modelx\largem & 81.7 / 81.1 & 76.0 & \textbf{81.7} & 74.0 & \textbf{82.1} & 68.5 & 96.1 / 94.6 & 77.1 / 36.3 & 77.0 \\
        &\docmodel & 80.2 / 79.6 & 77.0 & 78.8 & 76.2 & 79.8 & 63.6 & \textbf{97.3 / 96.4} & 74.6 / 32.1 & 75.7 \\
        &\sentmodel & 80.7 / 80.2 & 77.0 & 79.8 & 79.1 & 80.8 & 70.4 & 94.6 / 93.7 & 76.9 / 36.1 & 76.8\\
        &\modelx$_\text{410M}$ & 81.5 / 80.9 & 80.0 & \textbf{81.7} & \textbf{79.4} & 81.9 & 69.0 & 93.2 / \textbf{96.4} & 76.2 / 35.5 & 78.0\\
        &\modelx$_\text{515M}$ & \textbf{82.3 / 81.7} & \textbf{85.0} & \textbf{81.7} & 79.1 & 81.3 & 69.4 & 95.0 / \textbf{96.4} & \textbf{77.2} / 35.0 &\textbf{78.8}\\
        \midrule\midrule
        \multicolumn{5}{l}{\textit{Pretrained on larger corpora}}\\[0.02in]
        &T5\base & 76.2 / 75.4 & 73.0 & 79.8 & 78.3 & 80.8 & 67.9& 94.8 / 92.9 & 76.4 / 40.0 & 76.0 \\
        &T5\largem & 85.7 / 85.0 & 78.0 & \textbf{84.6} & 84.8 & 84.3 & 71.6 & 96.4 / 98.2 & 80.9 / 46.6 & 81.2 \\
        &BART\largem & 88.3 / 87.8 & 60.0 & 65.4 & 84.5 & 84.3 & 69.0 & 90.5 / 92.9 & 81.8 / 48.0 & 76.0 \\
        &RoBERTa\largem & 89.0 / 88.4 & \textbf{90.0} & 63.5 & 87.0 & \textbf{86.1} & \textbf{72.6} & 96.1 / 94.6 & \textbf{84.4 / 52.9} & 81.5 \\
        &\modelx\roberta & \textbf{89.6 / 89.0} & 82.0 & 83.7 & \textbf{87.7} & 84.7 & 71.2& \textbf{98.7 / 98.2} & 82.4 / 50.1 & \textbf{82.9}\\
        \bottomrule
    \end{tabular}
    }
\end{table*}

\section{Experiments}
\label{sec:experiment}

We now describe our pretraining setup and the evaluation of downstream tasks.

% We conduct experiments in two settings. In \Cref{subsec:single_obj}, we pretrain \model with only the short span objective and evaluate it on the NLU tasks. 
%We show that pretraining with autoregressive blank filling, combined with the new formulation of classification tasks, can outperform finetuning bidirectional encoders with linear classifiers. 
% In \Cref{subsec:multi_obj}, we explore multi-task pretraining as discussed in \Cref{subsec:multitask}. 
%We show that \model can effectively share model parameters for different tasks.

\subsection{Pretraining Setup}
\label{subsec:data}
For a fair comparison with BERT~\cite{devlinBERT2019}, we use BooksCorpus~\cite{BookCorpusZhu5} and English Wikipedia as our pretraining data. We use the uncased wordpiece tokenizer of BERT with 30k vocabulary. 
% Each input text is a document sampled from the corpus, with a start-of-sequence token $[\text{CLS}]$ prepended at the beginning\footnote{This token is not necessary for \model and only added for ablation study.} and an end-of-sequence token $[\text{EOS}]$ appended at the end. If the document is longer than the maximum sequence length of the transformer model, we randomly sample a continuous segment of the maximum length. 
% to get the sequence representations in downstream tasks for comparison with blank-filling finetuning.
We train \modelx\base and \modelx\largem with the same architectures as BERT\base and BERT\largem, containing 110M and 340M parameters respectively. 

For multi-task pretraining, we train two Large-sized models with a mixture of the blank infilling objective and the document-level or sentence-level objective, denoted as \docmodel and \sentmodel.  Additionally, we train two larger \model models of 410M (30 layers, hidden size 1024, and 16 attention heads) and 515M (30 layers, hidden size 1152, and 18 attention heads) parameters with document-level multi-task pretraining, denoted as \modelx$_\text{410M}$ and \modelx$_\text{515M}$.
% The output matrix of the softmax prediction is tied with the input embedding matrix.

% We use the AdamW optimizer~\cite{AdamW:LoshchilovH19} and set  $\beta_1=0.9,\beta_2=0.98,\epsilon=1e{-6}$. We adopt a linear warm-up of the learning rate over the first $8000$ steps and a cosine decay after that. The peak learning rate is set to $4e{-4}$ for \modelx\base and $2e{-4}$ for \modelx\largem. The weight decay is $0.1$ and the dropout rate is $0.1$.

To compare with SOTA models, we also train a Large-sized model with the same data, tokenization, and hyperparameters as RoBERTa~\cite{RoBERTa}, denoted as \modelx\roberta.
% \footnote{The STORIES dataset~\cite{Stories2019} used by RoBERTa is no longer available. Therefore, we replace OpenWebText (38GB)~\cite{OpenWebText2019} with OpenWebText2 (66GB)~\cite{pile}. The whole corpus totals 158GB of uncompressed text, close in size to RoBERTa's 160GB corpus.}
Due to resource limitations, we only pretrain the model for 250,000 steps, which are half of RoBERTa and BART's training steps and close to T5 in the number of trained tokens. 
More experiment details can be found in \Cref{sec:appendix_pretrain}.

\begin{table*}
    \caption{Results of abstractive summarization on the CNN/DailyMail and XSum test sets.}
    \label{tab:cnndm_xsum}
    \centering
    % \vspace{-0.08in}
    \small
    \resizebox{0.9\textwidth}{!}{
    \begin{tabular}{lcccccc}
        \toprule
        \multirowthead{2}{Model}& \multicolumn{3}{c}{CNN/DailyMail} & \multicolumn{3}{c}{XSum}\\
        & RG-1 & RG-2 & RG-L & RG-1 & RG-2 & RG-L\\
        \midrule
        BERTSumAbs~\cite{BERTSum2019} & 41.7 & 19.4 & 38.8 & 38.8 & 16.3 & 31.2 \\
        UniLMv2\base~\cite{UniLMv220} & 43.2 & 20.4 & 40.1 & 44.0 & 21.1 & 36.1 \\
        T5\largem~\cite{raffelT52020} & 42.5 & 20.7 & 39.8 & 40.9 & 17.3 & 33.0 \\
        BART\largem~\cite{lewisBART2019} & \textbf{44.2} & \textbf{21.3} & \textbf{40.9} & 45.1 & 22.3 & \textbf{37.3} \\
        \midrule
        \modelx\roberta & 43.8 & 21.0 & 40.5 & \textbf{45.5} & \textbf{23.5} & \textbf{37.3}\\
        \bottomrule
    \end{tabular}
    }
\end{table*}

\subsection{SuperGLUE}
\label{subsec:single_obj}

To evaluate our pretrained \model models, we conduct experiments on the SuperGLUE benchmark~\cite{SuperGLUE2019} and report the standard metrics. SuperGLUE consists of 8 challenging NLU tasks.
We reformulate the classification tasks as blank infilling with human-crafted cloze questions, following PET~\cite{PET2:2009-07118}. Then we finetune the pretrained \model models on each task as described in \Cref{subsec:finetune}. The cloze questions and other details can be found in \Cref{subsec:appendix_superglue}.
% For finetuning, we also use the AdamW optimizer with peak learning rate $1e{-5}$, warm-up over the first $6\%$ training steps and then a linear decay. For small datasets (COPA, WSC, CB, RTE), we finetune \model for 20 epochs. For larger datasets, we reduce the number of training epochs (10 for WiC, BoolQ, MultiRC; 5 for ReCoRD) as the model converges earlier.

For a fair comparison with \modelx\base and \modelx\largem, we choose BERT\base and BERT\largem as our baselines, which are pretrained on the same corpus and for a similar amount of time. We report the performance of standard finetuning (i.e. classification on the [CLS] token representation). The performance of BERT with cloze questions is reported in \Cref{subsec:ablation}. To compare with \modelx\roberta, we choose T5, BART\largem, and RoBERTa\largem as our baselines. T5 has no direct match in the number of parameters for BERT\largem, so we present the results of both T5\base (220M parameters) and T5\largem (770M parameters). All the other baselines are of similar size to BERT\largem.

\Cref{tab:superglue} shows the results. With the same amount of training data, \model consistently outperforms BERT on most tasks with either base or large architecture. The only exception is WiC (word sense disambiguation). On average, \modelx\base scores 4.6\% higher than BERT\base, and \modelx\largem scores 5.0\% higher than BERT\largem. It clearly demonstrates the advantage of our method in NLU tasks. In the setting of RoBERTa\largem, \modelx\roberta can still achieve improvements over the baselines, but with a smaller margin. Specifically, \modelx\roberta outperforms T5\largem but is only half its size. We also find that BART does not perform well on the challenging SuperGLUE benchmark. We conjecture this can be attributed to the low parameter efficiency of the encoder-decoder architecture and the denoising sequence-to-sequence objective. 

\subsection{Multi-Task Pretraining}
\label{subsec:multi_obj}
Then we evaluate the \model's performance in a multi-task setting (\Cref{subsec:objective}). Within one training batch, we sample short spans and longer spans (document-level or sentence-level) with equal chances. We evaluate the multi-task model for NLU, seq2seq, blank infilling, and zero-shot language modeling.

\textbf{SuperGLUE.}
For NLU tasks, we evaluate models on the SuperGLUE benchmark. The results are also shown in \Cref{tab:superglue}. We observe that with multi-task pretraining, \docmodel and \sentmodel perform slightly worse than \modelx\largem, but still outperform BERT\largem and UniLM\largem. Among multi-task models, \sentmodel outperforms \docmodel by 1.1\% on average. Increasing \docmodel's parameters to 410M (1.25$\times$BERT\largem) leads to better performance than \modelx\largem. \model with 515M parameters (1.5$\times$BERT\largem) can perform even better.

\begin{table}[t]
%  \begin{minipage}[t]{0.49\textwidth}
  \centering
  \small
    \caption{Results on Gigaword summarization.}
     \label{tab:gigaword}
    \resizebox{0.8\linewidth}{!}{
    \begin{tabular}{lccc}
        \toprule
        % & \multicolumn{3}{c}{Gigaword}\\
        Model & RG-1 & RG-2 & RG-L\\
        \midrule
        MASS & 37.7 & 18.5 & 34.9 \\
        UniLM\largem & 38.5 & 19.5 & 35.8 \\
        \midrule
        \modelx\largem & 38.6 & 19.7 & 36.0 \\
        \modelx\largemulti & 38.5 & 19.4 & 35.8\\
        \modelx$_\text{Sent}$ & 38.9 & 20.0 & \textbf{36.3}\\
        \modelx$_\text{410M}$ & \textbf{38.9} & \textbf{20.0} & 36.2 \\
        % \model$_\text{515M}$ (multi-task) \\
        \bottomrule
    \end{tabular}
    }
%   \end{minipage}
%   \hfill
    \vspace{0.2in}
   \centering
    \caption{Results on SQuAD question generation.}
        \label{tab:squad_qg}
    \resizebox{0.88\linewidth}{!}{
     \begin{tabular}{lccc}
        \toprule
        Model &  BLEU-4 & MTR & RG-L\\
        \midrule
        SemQG & 18.4 & 22.7 & 46.7\\
        UniLM\largem & 22.1 & 25.1 & \textbf{51.1}\\
        \midrule
        \modelx\largem  & 22.4 & 25.2 & 50.4\\
        \modelx\largemulti & 22.3 & 25.0 & 50.2\\
        \modelx$_\text{Sent}$ & 22.6 & 25.4 & 50.4 \\
        \modelx$_\text{410M}$ & \textbf{22.9} & \textbf{25.6} & 50.5\\
        \bottomrule
    \end{tabular}
    }
%    \end{minipage}
    \vspace{0.2in}
    \caption{BLEU scores on Yahoo text infilling. $^\dagger$ indicates the results from \cite{BLM2020}.}
    \label{tab:yahoo}
    \resizebox{\linewidth}{!}{
    \begin{tabular}{lccccc}
        \toprule
        Mask ratio         & 10\%&	20\%&	30\%&	40\%&	50\% \\\midrule
        BERT$^\dagger$     &	82.8&	66.3&	50.3&	37.4&	26.2 \\
        BLM$^\dagger$     &    86.5&	73.2&	59.6&	46.8&	34.8 \\
        \modelx\largem      &	\textbf{87.8}&	\textbf{76.7}&	\textbf{64.2}&	\textbf{48.9}&	\textbf{38.7} \\
        \modelx\largemulti  &	87.5&	76.0&	63.2&	47.9&	37.6 \\
        \bottomrule
    \end{tabular}
    }
\end{table}

\textbf{Sequence-to-Sequence.}
\label{sec:seq2seq}
Considering the available baseline results, we use the Gigaword dataset~\cite{GigawordRushCW15} for abstractive summarization and the SQuAD 1.1 dataset~\cite{SQuAD:RajpurkarZLL16} for question generation~\cite{NQG:du2017} as the benchmarks for models pretrained on BookCorpus and Wikipedia. Additionally, we use the CNN/DailyMail~\cite{GetPointSummarization2017} and XSum~\cite{XSum2018} datasets for abstractive summarization as the benchmarks for models pretrained on larger corpora. 
% We use abstractive summarization and question generation as the evaluation tasks. Abstractive summarization aims to produce a concise and fluent summary conveying the key information in the input text. 
% We fine-tune \model on the training set for 4 epochs with AdamW optimizer.
% The learning rate has a peak value of $3e-5$, warm-up over the 6\% training steps and a linear decay. We also use label smoothing with rate 0.1~\cite{LabelSmooth}. The maximum document length is 192 and the maximum summary length is 32. 
% During decoding, we use beam search with beam size of 5 and remove repeated trigrams. We tweak the value of length penalty on the development set. 
% We use MASS~\cite{songMASS2019} and UniLM~\cite{UniLM19} as our baselines since BART~\cite{lewisBART2019} and PALM~\cite{PALM2020} are trained with much more data and steps.
% Question generation aims to generate the corresponding question given an input passage and an answer. This was first proposed by \cite{NQG:du2017} to enhance the reading comprehension models. We use  and follow the dataset split of \cite{NQG:du2017}. 
% The optimizer hyperparameters are the same as those of abstractive summarization. The maximum passage length is 464 and the maximum question length is 48. 
% During decoding, we use beam search with beam size 5 and tweak the value of length penalty on the development set. We use SemQG~\cite{SemQG:ZhangB19} and UniLM as our baselines.

The results for models trained on BookCorpus and Wikipedia are shown in \Cref{tab:gigaword,tab:squad_qg}. We observe that \modelx\largem can achieve performance matching the other pretraining models on the two generation tasks.
\sentmodel can perform better than \largemodel, while
\docmodel performs slightly worse than \modelx\largem. This indicates that the document-level objective, which teaches the model to extend the given contexts, is less helpful to conditional generation, which aims to extract useful information from the context. Increasing \docmodel's parameters to 410M leads to the best performance on both tasks.
The results for models trained on larger corpora are shown in \Cref{tab:cnndm_xsum}. \modelx\roberta can achieve performance matching the seq2seq BART model, and outperform T5 and UniLMv2.

\textbf{Text Infilling.}
Text infilling is the task of predicting missing spans of text which are consistent with the surrounding context \cite{TextInfill2019,Fillin2020,BLM2020}. \model is trained with an autoregressive blank infilling objective, thus can straightforwardly solve this task. We evaluate \model on the Yahoo Answers dataset~\cite{YahooAnswers} and compare it with Blank Language Model (BLM) \cite{BLM2020}, which is a specifically designed model for text infilling. From the results in \Cref{tab:yahoo}, \model outperforms previous methods by large margins (1.3 to 3.9 BLEU) and achieves the state-of-the-art result on this dataset. We notice that \modelx\largemulti slightly underperforms \modelx\largem, which is consistent with our observations in the seq2seq experiments.
% For each document in the dataset, we randomly mask a given ratio $r\in \{10\% \cdots 50\% \}$ of its tokens and contiguous masked tokens are collapsed into a single blank. 
% We finetune \model to autoregressively generate the missing spans and evaluate the generation's quality by its BLEU score against the original document. 

% \begin{table}[!h]
%     \caption{BLEU scores on Yahoo text infilling. $^\dagger$ indicates the results taken from \cite{BLM2020}.}
%     \label{tab:yahoo}
%     \centering
%     \begin{tabular}{lccccc}
%         \toprule
%         Mask ratio         & 10\%&	20\%&	30\%&	40\%&	50\% \\\midrule
%         BERT$^\dagger$     &	82.8&	66.3&	50.3&	37.4&	26.2 \\
%         BLM$^\dagger$     &    86.5&	73.2&	59.6&	46.8&	34.8 \\
%         \modelx\largem      &	\textbf{87.8}&	\textbf{76.7}&	\textbf{64.2}&	\textbf{48.9}&	\textbf{38.7} \\
%         \modelx\largemulti  &	87.5&	76.0&	63.2&	47.9&	37.6 \\
%         \bottomrule
%     \end{tabular}
% \end{table}

\textbf{Language Modeling.}
% \begin{table}[!h]
%     \caption{Zero-shot language modeling results.}
%     \label{tab:language_model}
%     \centering
%     \begin{tabular}{l|r|r|r|rr}
%         \toprule
%         Model & \makecell{Lambada \\Acc. (uni)} & \makecell{BookWiki\\ PPL (uni)} & \makecell{Lambada \\Acc. (bi)} & \makecell{BookWiki\\ PPL (bi)}\\
%         \midrule
%         \modelx\largem & 0.0 & $>100$ & 10.6 & $>100$\\
%         \modelx\largemulti & 47.4 & 15.1 & 48.5 & 14.9 \\
%         \hspace{1em}-- 2d position  & 45.8 & 15.1 & 47.3 & 15.0\\
%         \model$_\text{410M,\multi}$  & 49.5 & 14.5 & \textbf{53.5} & \textbf{14.3}\\
%         \model$_\text{515M,\multi}$  & \textbf{50.4} & \textbf{13.9} & \textbf{54.9} & \textbf{13.7}\\
%         \midrule
%         GPT\largem & 50.1 & 14.4 & -- & -- \\
%         \bottomrule
%     \end{tabular}
% \end{table}
Most language modeling datasets such as WikiText103 are constructed from Wikipedia documents, which our pretraining dataset already contains.
% For fair comparison with BERT, we cannot remove Wikipedia from the pretraining dataset.
Therefore, we evaluate the language modeling perplexity on a held-out test set of our pretraining dataset, which contains about 20M tokens, denoted as BookWiki. We also evaluate GLM on the LAMBADA dataset~\cite{LAMBADADataset2016}, which tests the ability of systems to model long-range dependencies in text. The task is to predict the final word of a passage. As the baseline, we train a GPT\largem model~\cite{GPT2-2018,GPT3-2020} with the same data and tokenization as \modelx\largem. 

\begin{figure}
    \centering
    \includegraphics[width=0.8\linewidth]{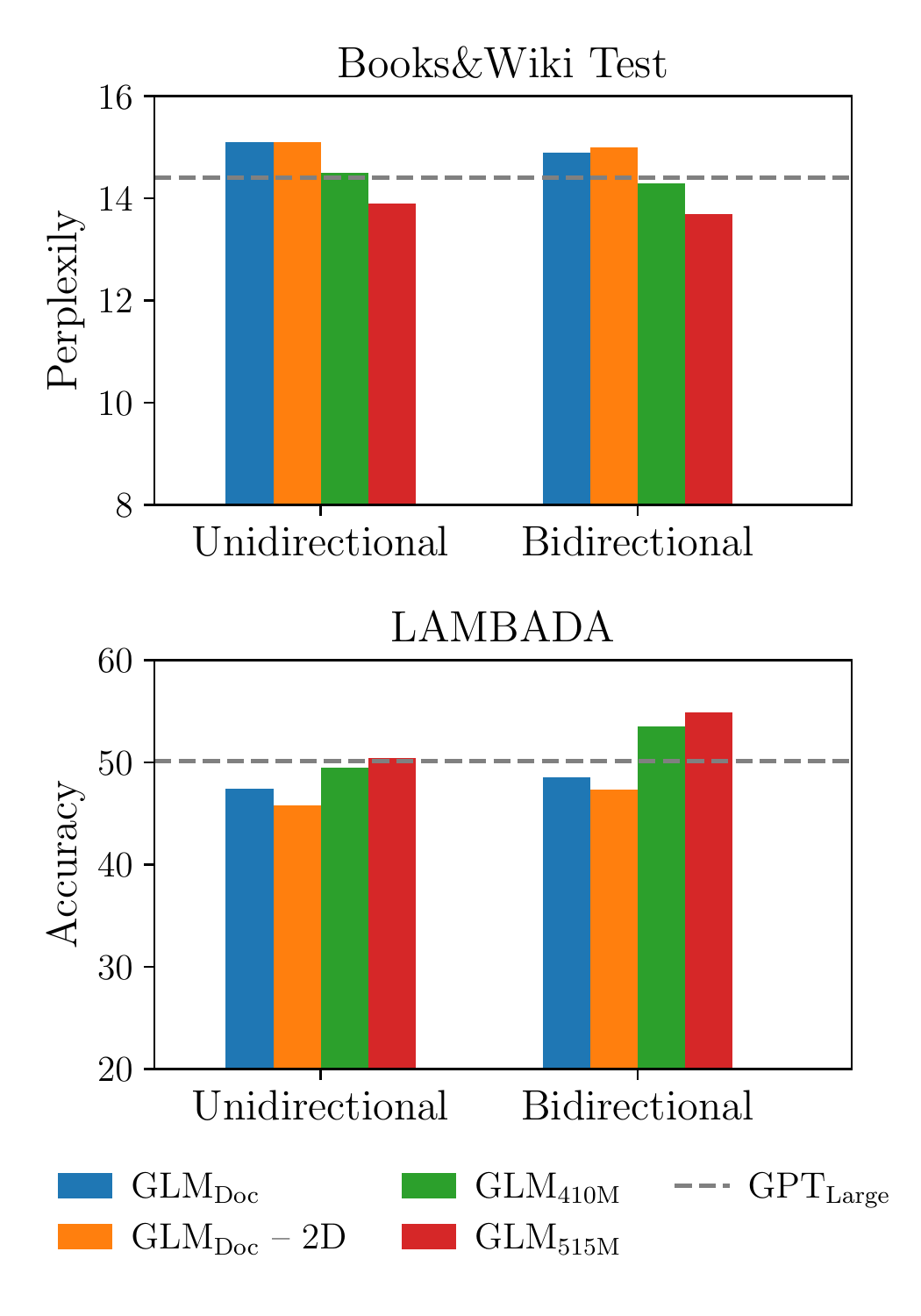}
    \vspace{-0.1in}
    \caption{Zero-shot language modeling results.}
    % \resizebox{\textwidth}{!}{
    %     \begin{tabular}{l|r|r}
    %         \toprule
    %         Model & \makecell{Lambada \\(Accuracy)} & \makecell{BookWiki\\ (Perplexity)}\\
    %         \midrule
    %         \model\largem (uni) & 0.0 & $>100$\\
    %         \model\largem (multi-task,uni) & 47.4 & 15.1 \\
    %         \hspace{1em}-- 2d positional encoding & 45.8 & 15.1\\
    %         \model$_\text{410M}$  (multi-task,uni) & 49.5 & 14.5\\
    %         \model$_\text{515M}$  (multi-task,uni) & \textbf{50.4} & \textbf{13.9}\\
    %         \midrule
    %         \model\largem (bi) & 10.6 & $>100$\\
    %         \model\largem (multi-task,bi) & 48.5 & 14.9\\
    %         \hspace{1em}-- 2d positional encoding & 47.3 & 15.0\\
    %         \model$_\text{410M}$  (multi-task,bi) & \textbf{53.5} & \textbf{14.3}\\
    %         \model$_\text{515M}$  (multi-task,bi) & \textbf{54.9} & \textbf{13.7}\\
    %         \midrule
    %         GPT\largem (uni) & 50.1 & 14.4 \\
    %         \bottomrule
    %     \end{tabular}
    % }
    \label{tab:language_model}
\end{figure}

The results are shown in \Cref{tab:language_model}. All the models are evaluated in the zero-shot setting. Since \model learns the bidirectional attention, we also evaluate \model under the setting in which the contexts are encoded with bidirectional attention. Without generative objective during pretraining, \modelx\largem cannot complete the language modeling tasks, with perplexity larger than 100. With the same amount of parameters, \modelx\largemulti performs worse than GPT\largem. This is expected since \docmodel also optimizes the blank infilling objective. Increasing the model's parameters to 410M (1.25$\times$ of GPT\largem) leads to a performance close to GPT\largem. \modelx$_\text{515M}$ (1.5$\times$ of GPT\largem) can further outperform GPT\largem. With the same amount of parameters, encoding the context with bidirectional attention can improve the performance of language modeling. Under this setting, \modelx$_\text{410M}$ outperforms GPT\largem. This is the advantage of \model over unidirectional GPT. We also study the contribution of 2D positional encoding to long text generation. We find that removing the 2D positional encoding leads to lower accuracy and higher perplexity in language modeling.

% \begin{table}[!t]
% \centering
%     \centering
%     % \small
%     \caption{BLEU scores on Yahoo text infilling. $^\dagger$ indicates the results from \cite{BLM2020}.}
%     % \vspace{-0.08in}
%     \begin{tabular}{lccccc}
%         \toprule
%         Mask ratio         & 10\%&	20\%&	30\%&	40\%&	50\% \\\midrule
%         BERT$^\dagger$     &	82.8&	66.3&	50.3&	37.4&	26.2 \\
%         BLM$^\dagger$     &    86.5&	73.2&	59.6&	46.8&	34.8 \\
%         \modelx\largem      &	\textbf{87.8}&	\textbf{76.7}&	\textbf{64.2}&	\textbf{48.9}&	\textbf{38.7} \\
%         \modelx\largemulti  &	87.5&	76.0&	63.2&	47.9&	37.6 \\
%         \bottomrule
%     \end{tabular}
%     % \vspace{0.05in}
%     \label{tab:yahoo}
% \end{table}

\begin{table*}
    \caption{Ablation study on the SuperGLUE dev set. (T5 $\approx$ GLM -- shuffle spans + sentinel tokens.)}
    \label{tab:classifier}
    % \vspace{-0.08in}
    \centering
    \small
    \begin{tabular}{l*{10}{r}}
        \toprule
        Model &  \makecell{ReCoRD\\F1/Acc.} & \makecell{COPA\\Acc.} & \makecell{WSC\\Acc.} & \makecell{RTE\\Acc.} & \makecell{BoolQ\\Acc.} &  \makecell{WiC\\Acc.} & \makecell{CB\\F1/Acc.}  & \makecell{MultiRC\\F1a/EM} & Avg \\
        \midrule
        % BERT\base (cloze) & 61.3/60.6 & 70.0 & 62.5 & 69.0 & 73.7 & \textbf{66.8} & 74.9/82.1 & 63.8/16.3 & 65.2 \\
        % \modelx\base (classifier) & 35.2/34.4 & 58.0 & 63.5 & 61.4 &	74.9 & 58.3 & \textbf{94.8/92.9} & 46.3/$\ \ $5.4 & 58.8\\
        % \modelx\base & \textbf{73.5/72.8} & \textbf{71.0} & \textbf{72.1} & \textbf{71.2} & \textbf{77.0} & 64.7 & 89.5/85.7 & \textbf{72.1/26.1} & \textbf{70.7}\\
        % \midrule
        BERT\largem & 76.3 / 75.6 & 69.0 & 64.4 & 73.6 & 80.1 & \textbf{71.0} & 94.8 / 92.9 & 71.9 / 24.1 & 72.0 \\
        BERT\largem (reproduced) & \textbf{82.1 / 81.5} & 63.0 & 63.5 & 72.2 & 80.8 & 68.7 & 80.9 / 85.7 & 77.0 / 35.2 & 71.2 \\
        BERT\largem (cloze) & 70.0 / 69.4 & \textbf{80.0} & 76.0 & 72.6 & 78.1 & 70.5 & 93.5 / 91.1 & 70.0 / 23.1 & 73.2 \\
        \modelx\largem & 81.7 / 81.1 & 76.0 & \textbf{81.7} & 74.0 & \textbf{82.1} & 68.5 & \textbf{96.1 / 94.6} & 77.1 / 36.3 & \textbf{77.0}\\
        \hspace{1em}-- cloze finetune & 81.3 / 80.6 & 62.0 & 63.5 & 66.8 & 80.5 & 65.0 & 89.2 / 91.1 & 72.3 / 27.9 & 70.0\\
        \hspace{1em}-- shuffle spans & 82.0 / 81.4 & 61.0 & 79.8 & 54.5 & 65.8 & 56.3 & 90.5 / 92.9 & 76.7 / 37.6 & 68.5\\
        \hspace{1em}+ sentinel tokens & 81.8 / 81.3 & 69.0 & 78.8 & \textbf{77.3} & 81.2 & 68.0 & 93.7 / 94.6 & \textbf{77.5 / 37.7} & 76.0\\
        \bottomrule
    \end{tabular}
\end{table*}

\textbf{Summary.} 
Above all, we conclude that \model effectively shares model parameters across natural language understanding and generation tasks, achieving better performance than a standalone BERT, encoder-decoder, or GPT model. % with fewer than two times its parameters.

\subsection{Ablation Study}
\label{subsec:ablation}

\Cref{tab:classifier} shows our ablation analysis for \model. First, to provide an apple-to-apple comparison with BERT, we train a BERT\largem model  with our implementation, data, and hyperparameters (row 2). The performance is slightly worse than the official BERT\largem and significantly worse than GLM\largem. It confirms the superiority of \model over Masked LM pretraining on NLU tasks. Second, 
% we perform an ablation study to understand the importance of autoregressive pretraining, and the formulation of classification tasks as blank infilling. 
we show the SuperGLUE performance of \model finetuned as sequence classifiers (row 5) and BERT with cloze-style finetuning (row 3). Compared to BERT with cloze-style finetuning, \model benefits from the autoregressive pretraining. Especially on ReCoRD and WSC, where the verbalizer consists of multiple tokens, \model consistently outperforms BERT. This demonstrates \model's advantage in handling variable-length blank. Another observation is that the cloze formulation is critical for \model's performance on NLU tasks. For the large model, cloze-style finetuning can improve the performance by 7 points. Finally, we compare \model variants with different pretraining designs to understand their importance. Row 6 shows that removing the span shuffling (always predicting the masked spans from left to right) leads to a severe performance drop on SuperGLUE. Row 7 uses different sentinel tokens instead of a single $[\text{MASK}]$ token to represent different masked spans. The model performs worse than the standard \model. We hypothesize that it wastes some modeling capacity to learn the different sentinel tokens which are not used in downstream tasks with only one blank. In \Cref{tab:language_model}, we show that removing the second dimension of 2D positional encoding hurts the performance of long text generation.

We note that T5 is pretrained with a similar blank infilling objective. \model differs in three aspects: (1) \model consists of a single encoder, (2) \model shuffles the masked spans, and (3) \model uses a single [MASK] instead of multiple sentinel tokens. While we cannot directly compare \model with T5 due to the differences in training data and the number of parameters, the results in \Cref{tab:superglue,tab:classifier} have demonstrated the advantage of \model.
\section{Related Work}
\textbf{Pretrained Language Models.} Pretraining large-scale language models significantly improves the performance of downstream tasks. There are three types of pretrained models. First, autoencoding models learn a bidirectional contextualized encoder for natural language understanding via denoising objectives~\cite{devlinBERT2019,joshiSpanBERT2020,YangXLNet2019,RoBERTa,ALBERT:lan2020,ELECTRA:clark2020}. 
% For example, BERT~\cite{devlinBERT2019} pretrains a large transformer model~\cite{Transformer2017} via masked language modeling to obtain contextualized word representations. 
% SpanBERT~\cite{joshiSpanBERT2020} masks continuous spans of tokens for improved span representations. 
Second, autoregressive models are trained with a left-to-right language modeling objective~\cite{GPT2018a,GPT2-2018,GPT3-2020}. 
% GPT~\cite{GPT2018a} shows that the representations learned by generative pretraining can also improve language understanding. 
% XLNet~\cite{YangXLNet2019} generalizes the autoregressive model with permutation language modeling to learn bidirectional attention for language understanding tasks. 
Third, encoder-decoder models are pretrained for sequence-to-sequence tasks~\cite{songMASS2019,lewisBART2019,PALM2020,PEGASUS:zhang2020}. 
% MASS~\cite{songMASS2019} maps an input text with continuous spans masked to the masked tokens.

Among encoder-decoder models, BART~\cite{lewisBART2019} conducts NLU tasks by feeding the same input into the encoder and decoder, and taking the final hidden states of the decoder. Instead, T5~\cite{raffelT52020} formulates most language tasks in the text-to-text framework. However, both models require more parameters to outperform autoencoding models such as RoBERTa~\cite{RoBERTa}.
UniLM~\cite{UniLM19,UniLMv220} unifies three pretraining models under the masked language modeling objective with different attention masks.

% PALM~\cite{PALM2020} is pretrained for generating coherent text from a given context and adds a BERT-based autoencoding objective to the encoder.

\textbf{NLU as Generation.} Previously, pretrained language models complete classification tasks for NLU with linear classifiers on the learned representations. GPT-2~\cite{GPT2-2018} and GPT-3~\cite{GPT3-2020} show that generative language models can complete NLU tasks such as question answering by directly predicting the correct answers without finetuning, given task instructions or a few labeled examples. However, generative models require much more parameters to work due to the limit of unidirectional attention. Recently, PET~\cite{PET:2001-07676,PET2:2009-07118} proposes to reformulate input examples as cloze questions with patterns similar to the pretraining corpus in the few-shot setting. It has been shown that combined with gradient-based finetuning, PET can achieve better performance in the few-shot setting than GPT-3 while requiring only 0.1\% of its parameters. Similarly, \citet{Augmented2020} and \citet{Augmented2021} convert structured prediction tasks, such as sequence tagging and relation extraction, to sequence generation tasks. 

\textbf{Blank Language Modeling.} \citet{Fillin2020} and \citet{BLM2020} also study blanking infilling models. Different from their work, we pre-train language models with blank infilling objectives and evaluate their performance in downstream NLU and generation tasks.
\section{Conclusions}
\model is a general pretraining framework for natural language understanding and generation. We show that the NLU tasks can be formulated as conditional generation tasks, and therefore solvable by autoregressive models. \model unifies the pretraining objectives for different tasks as autoregressive blank infilling, with mixed attention masks and the novel 2D position encodings. Empirically we show that \model outperforms previous methods for NLU tasks and can effectively share parameters for different tasks. 
% In the future, we hope to scale \model to larger Transformer models and more data, and examine its performance in more settings such as knowledge probing and few-shot learning.

\section*{Acknowledgements}
The work is supported by the NSFC for Distinguished Young Scholar(61825602), and Beijing Academy of Artificial Intelligence (BAAI).
% This document has been adapted
% by Steven Bethard, Ryan Cotterell and Rui Yan
% from the instructions for earlier ACL and NAACL proceedings, including those for 
% ACL 2019 by Douwe Kiela and Ivan Vuli\'{c},
% NAACL 2019 by Stephanie Lukin and Alla Roskovskaya, 
% ACL 2018 by Shay Cohen, Kevin Gimpel, and Wei Lu, 
% NAACL 2018 by Margaret Mitchell and Stephanie Lukin,
% Bib\TeX{} suggestions for (NA)ACL 2017/2018 from Jason Eisner,
% ACL 2017 by Dan Gildea and Min-Yen Kan, 
% NAACL 2017 by Margaret Mitchell, 
% ACL 2012 by Maggie Li and Michael White, 
% ACL 2010 by Jing-Shin Chang and Philipp Koehn, 
% ACL 2008 by Johanna D. Moore, Simone Teufel, James Allan, and Sadaoki Furui, 
% ACL 2005 by Hwee Tou Ng and Kemal Oflazer, 
% ACL 2002 by Eugene Charniak and Dekang Lin, 
% and earlier ACL and EACL formats written by several people, including
% John Chen, Henry S. Thompson and Donald Walker.
% Additional elements were taken from the formatting instructions of the \emph{International Joint Conference on Artificial Intelligence} and the \emph{Conference on Computer Vision and Pattern Recognition}.

\bibliography{main}
\bibliographystyle{acl_natbib}

\appendix
\section{Pretraining Setting}
\label{sec:appendix_pretrain}
\subsection{Datasets}
To train \modelx\base and \modelx\largem, we use BookCorpus~\cite{BookCorpusZhu5} and Wikipedia used by BERT~\cite{devlinBERT2019}.

To train \modelx\roberta, we follow the pretraining datasets of RoBERTa~\cite{RoBERTa}, which consist of BookCorups~\cite{BookCorpusZhu5},Wikipedia (16GB), CC-News (the English portion of the CommonCrawl News dataset\footnote{\url{https://commoncrawl.org/2016/10/news-dataset-available}} 76GB), OpenWebText (web content extracted from URLs shared on Reddit with at least three upvotes\cite{OpenWebText2019}, 38GB) and Stories (subset of CommonCrawl data filtered to match the story-like style of Winograd schemas~\cite{Stories2019}, 31GB). The Stories dataset is no longer publicly available\footnote{\url{https://github.com/tensorflow/models/tree/archive/research/lm_commonsense\#1-download-data-files}}. Therefore, we remove the Stories dataset and replace OpenWebText with OpenWebText2\footnote{\url{https://openwebtext2.readthedocs.io/en/latest}} (66GB). The CC-News dataset is not publicly available and we use the CC-News-en published by \cite{CCNewsEn2020}. All the datasets used total 158GB of uncompressed texts, close in size to RoBERTa's 160GB datasets.

\subsection{Hyperparameters}
\begin{table*}[t]
    \centering
    \caption{Hyperparameters for pretraining}
    \label{tab:hyper_pretrain}
    \begin{tabular}{lccc}
    \toprule
        Hyperparameters & \model\base & \model\largem & \model\roberta \\
    \midrule
        Number of Layers & 12 & 24 & 24 \\
        Hidden size & 768 & 1024 & 1024 \\
        FFN inner hidden size & 3072 & 4096 & 4096 \\
        Attention heads & 12 & 16 & 16 \\
        Attention head size & 64 & 64 & 64 \\
        Dropout & 0.1 & 0.1 & 0.1 \\
        Attention Dropout & 0.1 & 0.1 & 0.1 \\
        Warmup Steps & 6k & 8k & 30K \\
        Peak Learning Rate & 4e-4 & 2e-4 & 4e-4 \\
        Batch Size & 1024 & 1024 & 8192 \\
        Weight Decay & 0.1 & 0.1 & 0.01 \\
        Max Steps & 120k & 200k & 250k \\
        Learning Rate Decay & Cosine & Cosine & Cosine \\
        Adam $\epsilon$ & 1e-6 & 1e-6 & 1e-6 \\
        Adam $\beta_1$ & 0.9 & 0.9 & 0.9 \\
        Adam $\beta_2$ & 0.98 & 0.98 & 0.98 \\
        Gradient Clipping & 1.0 & 1.0 & 1.0 \\
    \bottomrule
    \end{tabular}
\end{table*}
The hyperparameters for \modelx\base and \modelx\largem are similar to those used by BERT. For trade-off of training speed and fair comparison with BERT (batch size 256 and 1,000,000 training steps), we use batch size of 1024 and 200,000 training steps for \modelx\largem. Since \modelx\base is smaller, we reduce the number of training steps to 120,000 to speed up pre-training. The hyperparameters for \docmodel and \sentmodel are the same as those of \largemodel. The hyperparameters except Transformer architecture for \modelx$_\text{410M}$ and \modelx$_\text{515M}$ are the same as those of \largemodel. The models are trained on 64 V100 GPUs for 200K steps with batch size of 1024 and maximum sequence length of 512, which takes about 2.5 days for \largemodel. 

To train \modelx\roberta, we follow most of the hyperparameters of RoBERTa. The main difference includes: (1) Due to resource limit, we only pre-train \model\roberta for 250,000 steps, which are half of RoBERTa and BART's training steps, and close to T5 in number of trained tokens. (2) We use cosine decay instead of linear decay for learning rate scheduling (3) We additionally apply gradient clipping with value 1.0.

The hyperparameters for all the pre-training settings are summarized in \Cref{tab:hyper_pretrain}.

\subsection{Implementation}
Our pretraining implementation is based on Megatron-LM~\cite{Megatron-LM} and DeepSpeed~\cite{DeepSpeed:Rasley}. We include our code in the supplementary material. Due to the size limit of supplementary material, we cannot include the pretrained models, but will make them public available in the future.

\section{Downstream Tasks}
\subsection{SuperGLUE}
\label{subsec:appendix_superglue}
The SuperGLUE benchmark consists of 8 NLU tasks. We formulate them as blank infilling tasks, following \cite{PET2:2009-07118}. Table~\ref{tab:superglue_cloze} shows the cloze questions and verbalizers we used in our experiments. For 3 tasks (ReCoRD, COPA, and WSC), the answer may consist of multiple tokens, and for the other 5 tasks, the answer is always a single token. 

When finetuning \model on the SuperGLUE tasks, we construct the input using the cloze questions in Table~\ref{tab:superglue_cloze} and replace the blank with a [MASK] token. Then we compute the score of generating each answer candidate. For the 5 single-token tasks, the score is defined to be the logit of the verbalizer token. For the 3 multi-token tasks, we use the sum of the log-probabilities of the verbalizer tokens. Thanks to the autoregressive blank infilling mechanism we proposed, we can obtain all the log-probabilities in one pass. Then we compute the cross entropy loss using the groundtruth label and update the model parameters.

For the baseline classifiers, we follow the standard practice to concatenate the input parts of each task (such as the premise and hypothesis for textual entailment, or the passage, question and answer for ReCORD and MultiRC) and add a classification layer on top of the [CLS] token representation. We also implemented cloze-style finetuning for the other pre-trained models, but the performance was usually similar to the standard classifier, as we shown in the ablation study. Models with blank-infilling objectives, such as T5 and our \model, benefits more from converting the NLU tasks into cloze questions. Thus for T5 and \model, we report the performance after such conversion in our main results.

\begin{table*}[]
    \centering
    \caption{Cloze questions and verbalizers for the 8 SuperGLUE tasks used in our experiments. $^*$~denotes the answer contains multiple tokens.}
    \label{tab:superglue_cloze}
    \resizebox{1.0\textwidth}{!}{
    \begin{tabular}{llp{7cm}p{3cm}}
    \toprule
    Dataset & Task & Cloze Question & Verbalizers  \\
    \midrule 
    ReCoRD$^*$ & Question answering & [passage $p$] [cloze question $q$] & Answer candidates \\
    COPA$^*$ & Causal reasoning & ``[choice $c_1$]'' or ``[choice $c_2$]''? [premise $p$], so \rule{0.4cm}{0.4pt}. & $c_1$ / $c_2$ \\
    WSC$^*$ & Coreference resolution & [sentence $s$] The pronoun `$*p*$' refers to \rule{0.4cm}{0.4pt}. & Noun $n$ \\
    RTE & Textual entailment & ``[hypothesis $h$]''? $\mid$ \rule{0.4cm}{0.4pt}, ``[premise $p$]'' & ``yes'' (entailment), ``no'' (not entailment)\\
    BoolQ & Question answering & [passage $p$]. Question: $q$? Answer: \rule{0.4cm}{0.4pt}. & ``yes'' / ``no''  \\
    WiC & Word sense disambiguation & ``[sentence $s_1$]'' / ``[sentence $s_2$]'' Similar sense of [word $w$]? \rule{0.4cm}{0.4pt}. & ``yes'' / ``no'' \\
    CB & Textual entailment & ``[hypothesis $h$]''? $\mid$ \rule{0.4cm}{0.4pt}, ``[premise $p$]'' & ``yes'' (entailment), ``no'' (contradiction), ``maybe'' (neutral) \\
    MultiRC & Question answering & [passage $p$]. Question: $q$? Is it [answer $a$]? \rule{0.4cm}{0.4pt}. & ``yes'' / ``no'' \\
    \bottomrule
    \end{tabular}
    }
\end{table*}

\subsection{Sequence-to-Sequence}
Fot the text summarization task, we use the dataset Gigaword~\cite{GigawordRushCW15} for model fine-tuning and evaluation. We finetune \modelx$_{\text{LARGE}}$ on the training set for 4 epochs with AdamW optimizer. The learning rate has a peak value of 3e-5, warm-up over the 6\% training steps and a linear decay. We also use label smoothing with rate 0.1~\cite{LabelSmooth}. The maximum document length is 192 and the maximum summary length is 32. During decoding, we use beam search with beam size of 5 and remove repeated trigrams. We tweak the value of length penalty on the development set. The evaluation metrics are the F1 scores of Rouge-1, Rouge-2, and Rouge-L~\cite{ROUGE:Lin2004} on the test set.

For the question generation task, we use the SQuAD 1.1 dataset~\cite{SQuAD:RajpurkarZLL16} and follow the dataset split of \cite{NQG:du2017}. The optimizer hyperparameters are the same as those of abstractive summarization. The maximum passage length is 464 and the maximum question length is 48. During decoding, we use beam search with beam size 5 and tweak the value of length penalty on the development set. The evaluation metrics are the scores of BLEU-1, BLEU-2, BLEU-3, BLEU-4~\cite{Bleu:papineni2002}, METEOR~\cite{Meteor:denkowski2014} and Rouge-L~\cite{ROUGE:Lin2004}.

Results of T5\largem on XSum are obtained by running the summarization script provided by Huggingface transformers\footnote{\url{https://github.com/huggingface/transformers/tree/master/examples/pytorch/summarization}}. All the other results of baselines on seq2seq tasks are obtained from the corresponding papers.

\subsection{Text Infilling}
We follow \cite{BLM2020} and evaluate text infilling performance on the Yahoo Answers dataset \cite{YahooAnswers}, which contains 100K/10K/10K documents for train/valid/test respectively. The average document length is 78 words. To construct the text infilling task, we randomly mask a given ratio $r\in \{10\% \cdots 50\% \}$ of each document's tokens and the contiguous masked tokens are collapsed into a single blank. We finetune \modelx\largem on the training set for 5 epochs with dynamic masking, i.e. the blanks are randomly generated at training time. Similar to the sequence-to-sequence experiments, we use an AdamW optimizer with a peak learning rate 1e-5 and 6\% warm-up linear scheduler. 

For comparison with previous work, we use the same test set constructed by \cite{BLM2020}. The evaluation metric is the BLEU score of the infilled text against the original document. We compare with two baselines: (1) BERT, which learns a left-to-right language model to generate the masked tokens on top of the blank representation, and (2) BLM proposed by \cite{BLM2020}, which can fill in the blank with arbitrary trajectories.

\subsection{Language Modeling}
We evaluate the model's ability of language modeling with perplexity on BookWiki and accuracy on the LAMBDA dataset~\cite{LAMBADADataset2016}. 

Perplexity is an evaluation criterion that has been well studied for language modeling. Perplexity is the exponentiation of the average cross entropy of a corpus.
\begin{equation}
    \text{PPL}=\exp(-\frac{1}{T}\sum_{t=1}^Tp(x_t|\myvec{x}_{<t}))
\end{equation}
where $\myvec{x}_{<t}=[x_0,\cdots,x_{t-1}]$. Since transformers can only operate on a window of fixed input size $w$, we cannot fully calculate $p(x_t|\myvec{x}_{< t})$ and can only calculate $p(x_t|\myvec{x}_{t-w:t-1})$. Even calculating this value for each token is prohibitively expensive, since we need to conduct $T$ evaluations of $w$-size contexts. To improve evaluation efficiency, we adopt \emph{overlapping evaluation}, where we advance the sliding windows by some overlap $o$ each time and only compute the cross entropy loss for the last $o$ tokens of the window. In our experiments we set $o=256$ for all the models.

LAMBDA is a cloze-style dataset to test the ability of long-range dependency modeling. Each example is a passage consisting of 4-5 sentences with the last word missing and the model is required to predict the last word of the passage. Since we use WordPiece tokenization, a word can be split into several subword units. We use teacher forcing and consider the prediction correct only when all the predicted tokens are correct. 

\section{Results on Other NLU Benchmarks}
GLUE~\cite{GLUEBenchmark2018} is another widely-used NLU benchmark, including single sentence tasks (e.g. sentiment analysis~\cite{SentimentBank2013}) and sentence pair tasks (e.g. text similarity~\cite{TextualSimilarity2017} and natural language inference~\cite{MNLI2018,RTE2005}). The benchmark is usually considered as less challenging than SuperGLUE. SQuAD~\cite{SQuAD:RajpurkarZLL16,SQuADKnowWhatYou2018} is an extractive question answering benchmark. We further compare \model with BERT on the two benchmarks.

The results on GLUE and SQuAD are shown in \Cref{tab:glue_result,tab:squad_result}. On the two benchmarks, GLM can still outperform BERT with the same amount of parameters, but with a smaller margin.
\begin{table*}
    \centering
    \caption{Results on the GLUE dev set.}
    \label{tab:glue_result}
    \begin{tabular}{l*{9}{r}}
        \toprule
        Model & MNLI & QNLI & QQP & RTE & SST-2 & MRPC & CoLA & STS-B & Avg\\
        \midrule
        BERT\largem & 86.6 & 92.3 & 91.3 & 73.6 & 93.2 & 88.0 & 60.6 & 90.0 & 84.4\\
        GLM\largem &  \textbf{86.7} & \textbf{92.8} & \textbf{91.5} & \textbf{74.0} & \textbf{93.5} & \textbf{90.0} & \textbf{61.4} & \textbf{90.7} & \textbf{85.1}\\
        \bottomrule
    \end{tabular}
\end{table*}

\begin{table}
    \caption{Results on the SQuAD v1.1/v2.0 dev sets.}
    \label{tab:squad_result}
    \centering
    \begin{tabular}{lrr}
        \toprule
        Model & \makecell{SQuAD v1.1\\EM/F1} & \makecell{SQuAD v2.0\\EM/F1}\\
        \midrule
        BERT\base & 80.8 / 88.5 & 73.7 / 76.3\\
        GLM\base & \textbf{81.5} / \textbf{88.6} & \textbf{74.7} / \textbf{77.8} \\
        \midrule
        BERT\largem & 84.1 / 90.9 & 79.0 / 81.8 \\
        GLM\largem & \textbf{85.4} / \textbf{91.6} & \textbf{80.3} / \textbf{83.3} \\
        \bottomrule
    \end{tabular}
\end{table}

\section{Text Generation Samples}
We show texts generated by \docmodel given unseen contexts randomly sampled from the test set. We use top-$k$ random sampling with $k=40$ for generation and set maximum sequence length to 512. Some of the texts are cut short.

\begin{example}
    The Wyoming State Legislature is the legislative branch of the U.S. State of Wyoming. 
    
    \textbf{\model}: It consists of a state senate and a house of representatives. As of 2019, state senators are Steve Kish, Joe L. Williams, and Linda S. Bollens. Members of the Wyoming State Legislature are elected from single-member districts representing the majority of the state. The current state senate members are: In recent years, there have been four changes to the senate. The most recent is the creation of a six-seat district that includes all or part of the following: In the 2009 elections, the state senate members were elected to six-year terms. The current state house members are: The Wyoming Constitution assigns certain powers to the governor. Most notably, the governor is president of the senate and governor. However, if the governor desires to appoint a member to the Wyoming state senate, a law authorizes the governor to do so. The governor of Wyoming holds no legislative power but has the power to veto lawmakers, which is not limited to the veto of laws. Under the wyoming state constitution, the governor can veto the actions of the other members of the wyoming house of representatives. The governor can also appoint members of the wyoming senate. In addition, the governor can appoint members of the Wyoming house of representatives. Wyoming's constitution provides that the governor can appoint a member of the wyoming state senate to the wyoming supreme court, and the chairman of the wyoming senate.
\end{example}

\begin{example}
    Jonathan Terry is a television and film actor. 
    
    \textbf{\model}: He was a voice actor for the ``X-Men" cartoon series. He was also a voice actor for ``the Simpsons" as well as ``the marvelous misadventures of superman. He also has voiced characters in ``the legend of korra" He has appeared on several television series, including ``the simpsons", ``the x-files", ``heroes" and ``the simpsons movie" as the character captain billy higgledy-pig, and the web series ``krusty mysteries" as the character Colonel Trungus. He wrote for ``the Guardian" newspaper. Jonathan Terry was born in London. Terry grew up in Surrey, England and attended the University of Sussex in the United Kingdom, graduating with a degree in english literature. He was a guest lecturer at King's College London, and then took two years of acting courses at the brit school of acting to prepare for his future career in the entertainment industry. Terry first appeared in the TV series ``the Simpsons" as the character captain Billy Higgledy-pig, but his character was only a one-time recurring character in the series' first six seasons. He later appeared as a regular for the show's final six seasons, and has been a frequent guest in the show since. He appeared in the first few episodes of ``" as the character major Jack Ryan. He has also appeared as part of the supporting cast of several episodes of ``the secret life of pets". He has also worked on ``the simpsons" TV show since ``the simpsons movie", most notably playing the roles of Captain Skeletor and the ghost of the same name. He plays characters in several films, including ``", ``", ``" and ``". He has appeared in music videos for the killers in 1993, the pretenders in 1995, and in the TV shows ``the royal" and ``the bill".
\end{example}

\begin{example}
    Corona was a station along the port Washington branch of the long island rail road in the Corona section of queens, New York City. It was one of two stations built by the flushing railroad in Corona, this one having been at Grand Avenue (later called National Avenue, now National Street ) and 45th Avenue. 
    
    \textbf{\model}: It had a brick entrance building next to the tracks, and one platform with 2 side platforms. The station burned down and was rebuilt in 1908. The station closed on september 15, 1927, with the train service transferred from Grand Avenue to 45th Avenue. There were plans to build a subway extension to Corona, but it was never built. In the 1980s, two stations were constructed on the line, Corona Road and Corona Park. The new Corona station opened in 1988, and the original Corona station was demolished.
\end{example}

\begin{example}
    Robert Lee Smith (born July 5, 1938) is a former American football cornerback in the national football league. He played for the Los Angeles Rams (1962–1965) and the Detroit Lions (1965–1966). He was also a kick and punt returner. 
    
    \textbf{\model}: In his four-year NFL career, he played in 33 games and started 14, registering 62 career interceptions. He completed his NFL career with five interceptions. Smith was born in La Canada Flintridge, Michigan, in 1938. He attended Bishop O' Dowd high school in Flintridge. Smith' s father was a pharmacist. He graduated from Michigan State University in 1958 with a degree in business administration. Smith played for the Michigan Wolverines football team from 1959 to 1963. In his two years as a reserve cornerback, he led the conference in interceptions with five. In 1962, he set the Wolverines' all-time interception record with 13, and was second overall in the 1962 season's Heisman Trophy voting. Smith also won the Dick Butkus award as the nation's outstanding linebacker. In 1961, the ``Los Angeles Times" wrote that Smith ``is an outstanding pass rusher, with an average of almost 100 yards per punt return." Smith was inducted into the university of Michigan athletic hall of honor in 1989 and the national football foundation hall of fame in 1991. He was elected to the Michigan sports hall of fame in 1995. Smith earned the honor because of his accomplishments prior to his NFL career. He was one of four Michigan players honored as first-overall selections in the 1964 NFL draft. The others were Joe Namath, Bill Nelsen, and Jerry Kramer. In 1966, the NFL gave players \$300,000 a season to play football. After his rookie season, he was not selected to play in the 1966 pro bowl. On January 13, 1966, the Rams traded smith to the Detroit Lions for Paul Hornung, and later that year he was traded to the Lions for Ray ``the Lion" Jones in exchange for Linebacker Jim ``the Hawk" Johnson. On September 10, 1968, he was traded back to Los Angeles for a second round pick in the 1970 draft. He was also traded to the St. Louis Cardinals for a second round pick in the 1970 draft. On June 2, 1970 he was cut by the Cardinals. On November 15, 1970, the Los Angeles Rams acquired Smith from the Lions in exchange for Linebacker Tony Harris. The Rams waived Smith during the September 1, 1972 offseason. Smith's number at Michigan State was \# 7 in 1969.
\end{example}

\end{document}